\documentclass{article}

\usepackage[preprint]{nips_2018}

\usepackage[utf8]{inputenc} 
\usepackage[T1]{fontenc}    
\usepackage[colorlinks=true,urlcolor=blue]{hyperref}       
\usepackage{url}            
\usepackage{booktabs}       
\usepackage{amsfonts}       
\usepackage{nicefrac}       
\usepackage{microtype}      
\usepackage{graphicx}
\title{Generating a Training Dataset for Land Cover Classification to Advance Global Development}

\author{
  Yoni Nachmany \\
  Radiant Earth Foundation\\
  Oakland, CA 94403 \\
  \texttt{yoni.nachmany@radiant.earth} \\
  \And
  Hamed Alemohammad \\
  Radiant Earth Foundation\\
  Oakland, CA 94403 \\
  \texttt{hamed@radiant.earth} \\  
}

\begin{document}

\maketitle

\begin{abstract}
  Semantic segmentation of land cover classes is fundamental for agricultural and economic development work, from sustainable forestry to urban planning, yet existing training datasets have significant limitations. To generate an open and comprehensive training library of high resolution Earth imagery and high quality land cover classifications, public Sentinel-2 data at 10 m spatial resolution was matched with accurate GlobeLand30 labels from 2010, which were filtered by agreement with an intermediary Sentinel-2 classification at 20 m produced during atmospheric correction. Scene-level classifications were predicted by Random Forests trained on valid reflectance data and the filtered labels, and achieved over 80\% model accuracy for a variety of locations. Further work is required to aggregate individual scene classifications for annual labels and to test the approach in more locations, before crowdsourcing human validation. The goal is to create a sustained community-wide effort to generate image labels not only for land cover, but also very specific images for major agriculture crops across the world and other thematic categories of interest to the global development community. 

\end{abstract}

\section{Introduction}

Advances in sensor technology, cloud computing, and machine learning (ML) continue to converge to accelerate innovation in the field of Earth observation (EO). In recent years, significant advancements have been made by the commercial sector in developing ML based algorithms for satellite imagery to extract intelligence on agricultural productivity, oil storage, urban structures, and maritime monitoring. These successful efforts from the commercial sector underscore the enormous potential of using ML to solve global development and humanitarian challenges. However, fundamental tools and technologies still need to be developed to drive further breakthroughs and to ensure that the Global Development Community (GDC) reaps the same benefits that the commercial marketplace is experiencing. 

Radiant Earth Foundation - a non-profit organization with a mission to improve discovery, access, delivery, and application of open geospatial resources in support of the GDC - proposes to advance critical insights in support of global development and humanitarian response through integrating and exploiting the latest in satellite data analytics and information technology. Radiant Earth Foundation is developing open source datasets of labeled satellite images, which will be hosted on \href{https://www.mlhub.earth/}{MLHub.Earth} with a Creative Commons license. These datasets will lead to a living open image library for ML and EO. Our goal is to create a sustained, community-wide effort to generate image labels that would enable major innovations and will drive new, more targeted and timely insights supporting progress in areas such as agriculture, food security, conservation, health, land rights, urban planning, water resources, and other areas relevant to global development and humanitarian response.

This paper focuses on developing an openly available dataset of global land cover (LC) labeled imagery from Sentinel-2 satellites at 10 m spatial resolution through Radiant Earth Foundation’s platform to enable fully-automated and dynamic LC classification algorithms. The approach for labelling these images uses a combination of machine learning and crowdsourcing to generate a human-verified training dataset. Existing training datasets for LC classification have limitations that do not support development of a global EO-based LC classification algorithm at fine spatial resolutions with high accuracy. These datasets are either generated for specific regions of the world (therefore, they lack geo-diversity) or are based on imagery that are not freely available at the global scale (therefore, they are not open source)~\citep{ISPRS2D, Campos-Taberner2016, Demir2018}. Moreover, in many cases, very few labeled images are available for a specific class within the dataset, which limits the performance of a ML algorithm to learn the particular features of that class.

\section{Data}
\subsection{Satellite Imagery}

Multispectral data from the constellation of Sentinel-2 satellites is publicly available through the European Space Agency (ESA). Atmospherically-corrected reflectance images are also regularly generated across Europe, but can be manually generated elsewhere. Therefore, initial experiments of this study are focused on ${100 \times 100}$ $km^2$ tiles across Europe representing a variety of LC conditions. The same approach will be attempted globally following the development of an atmospheric-correction pipeline. Geo-diversity~\citep{Shankar2017, Bollinger2018} is an important feature of the end dataset, and the approach will be adapted to ensure high performance around the world.

To produce annual LC labels, images were collected from July 2017 to July 2018 (Sentinel-2 has a revisit rate of 5 days). Images with more than 90\% cloud cover were filtered out. Our experiments included using only the four spectral bands of Sentinel-2 images with 10 m resolution (RGB and NIR, or near-infrared) as predictors, as well as combining those with the other six bands at 20 m (in red edge and SWIR, or short-wave infrared), bilinearly resampled to 10 m. Results showed that inclusion of the extra six bands improves classification of pixels with vegetation and snow.

\subsection{Reference Data}

\begin{figure}
  \centering
  \includegraphics[width=0.7\textwidth]{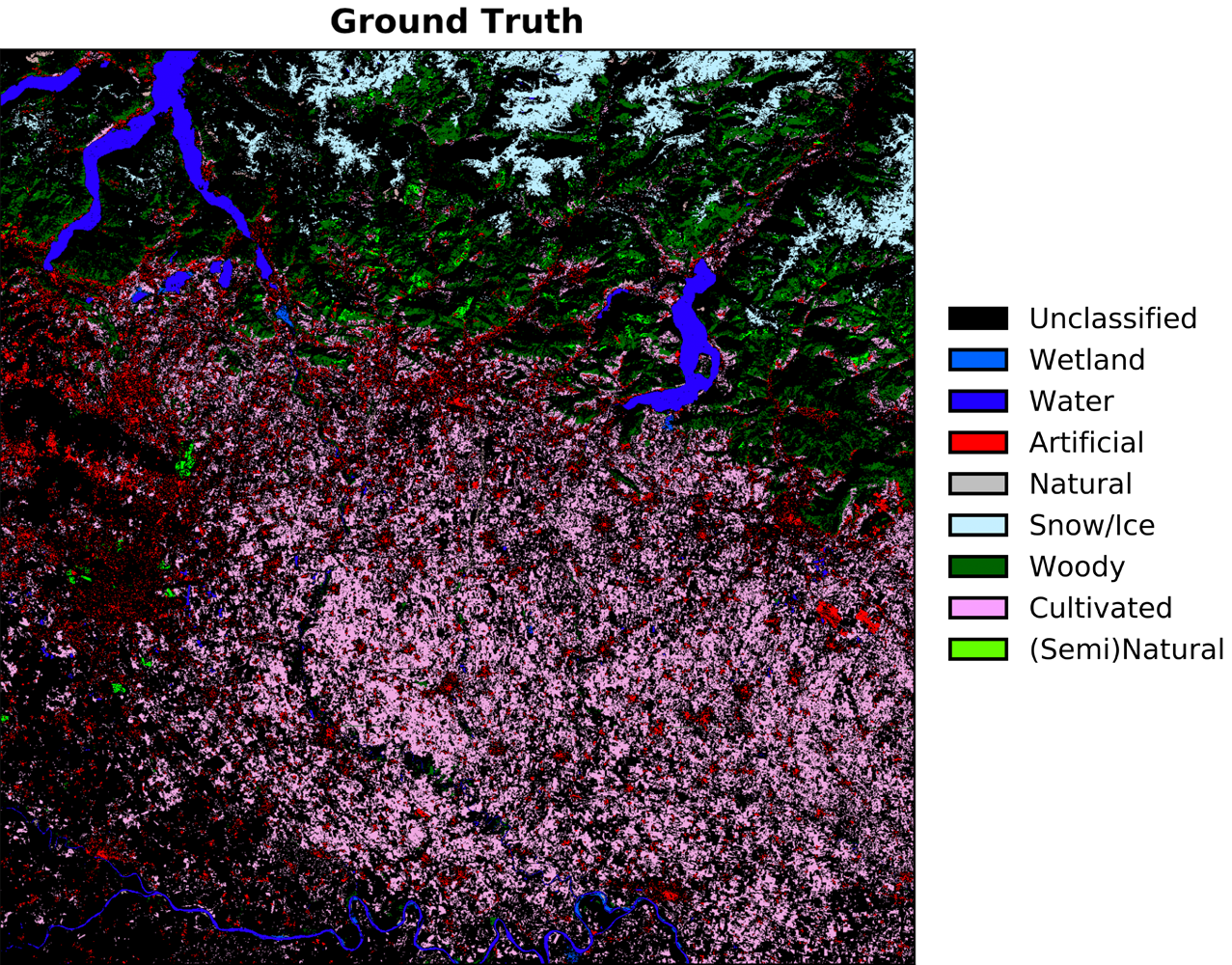}
  \caption{Example of ground truth labels obtained by mapping GLC labels and filtering by S2 labels}
  \label{GT}
\end{figure}

A major challenge for predicting LC labels at the 10 m spatial resolution of Sentinel-2 was preparing labeled training data from quality existing global datasets that are coarser and older. GlobeLand30 is the first 30 m resolution global LC dataset, with 10 classes and over 80\% accuracy for the year 2010~\citep{Chen2015}. First, GlobeLand30 labels were mapped to the last level of a hierarchical LC taxonomy~\citep{REFTaxonomy} developed by Radiant Earth Foundation’s Working Group on Machine Learning for Global Development~\citep{Alemohammad2018} and re-gridded using nearest neighbor interpolation to match Sentinel-2 data. Then, to better reflect Sentinel-2 imagery for the year starting in July 2017, GlobeLand30 labels were filtered by agreement with classes from Sentinel-2’s 20 m scene classifications~\citep{ESA2018}, which are produced in the process of Level-2A atmospheric correction and have been independently validated~\citep{Main-Knorn2017}. The filtered labels are used as ground truth labels for training [Figure~\ref{GT}].

\begin{figure}
  \centering
  \includegraphics[width=0.7\textwidth]{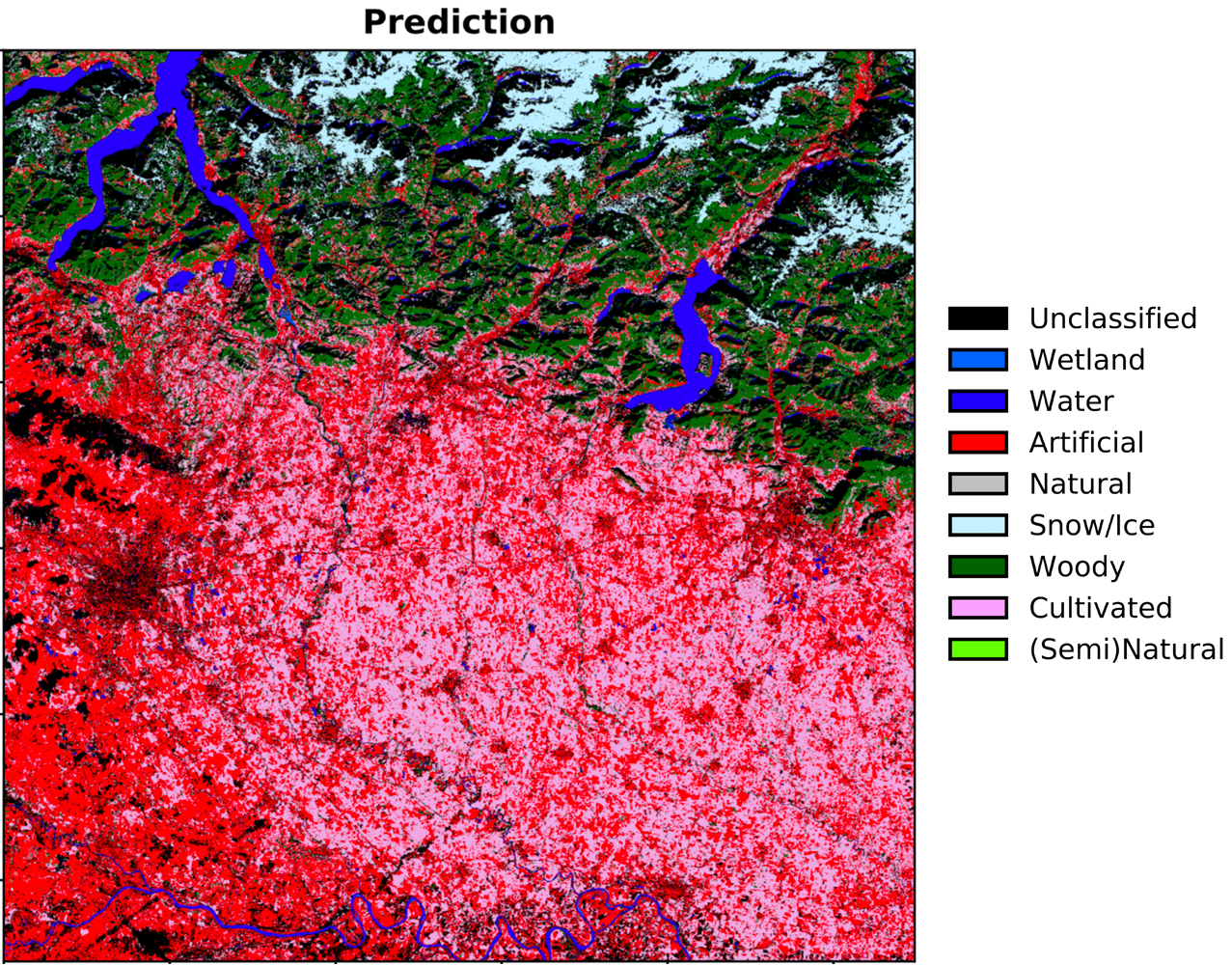}
  \caption{RF prediction; most of unclassified areas in ground truth data are filled with “woody vegetation“ and “artificial bare ground“}
  \label{prediction}
\end{figure}

\begin{figure}
  \centering
  \includegraphics[width=0.7\textwidth]{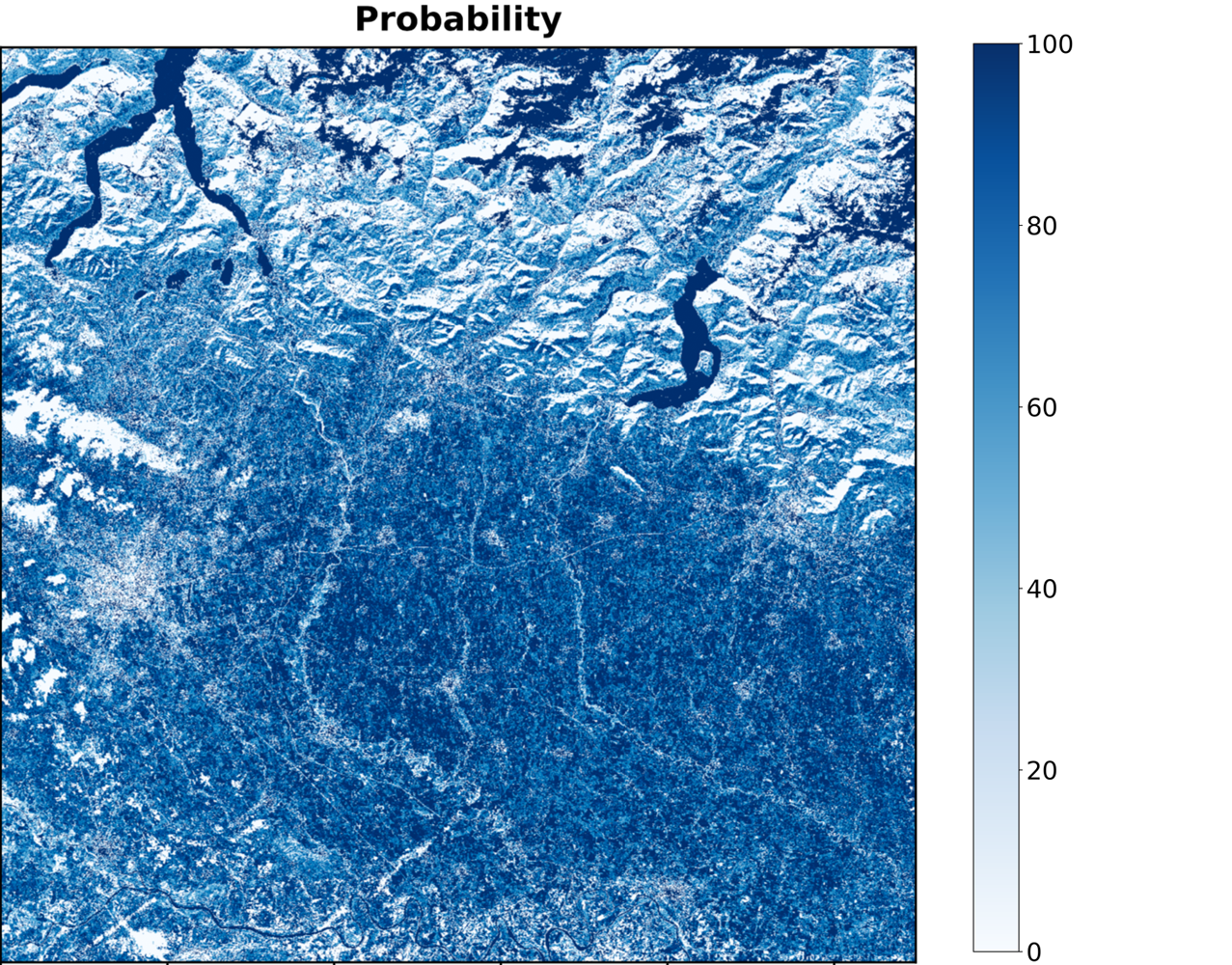}
  \caption{Prediction probabilities, very high for “water“ and “snow/ice“ and relatively high for other classes.}
  \label{probability}
\end{figure}

\section{Methodology}

For land cover classification on sub-meter resolution satellite imagery, like the ISPRS 2D Semantic Labeling Contest~\citep{ISPRS2D}, deep fully convolutional neural networks with encoder-decoder architectures have improved the state-of-the-art~\citep{Audebert2017}. However, for land cover classification on Sentinel-2 imagery at 10 m spatial resolution, experts recommend ensemble methods, which are widely adopted in practice~\citep{ML4GDAlgos2018}. ESA’s Sentinel-2 Global Land Cover (S2GLC) project selected a pixel-based, supervised approach using Random Forests “based on classification accuracy, preservation of class details and processing efficiency”~\citep{Malinowski2018}. S2GLC performed tile-wise training, with a large number of samples per class, and then aggregated predictions from individual scenes over a year for annual labels. Accordingly, processing is performed at the scene-level, on pixels with valid reflectance values in areas with <90\% average cloud cover (cloud confidence masks are provided during Level-2A processing). Random Forests are trained and tested on class-stratified samples of half the pixels in a scene, with one Sentinel-2 pixel at 10 m for each label pixel at 30 m. Predictions are made on all pixels marked with usable classes during Level-2A processing, including pixels labeled as ‘unclassified’ [Figure ~\ref{prediction}]. Random Forests also provide probabilities for predicted classes, which are written [Figure ~\ref{probability}] and can be used for aggregation of a time series of predictions for a given tile~\citep{Nowakowski2017}.

\begin{figure}
  \centering
  \includegraphics[width=0.65\textwidth]{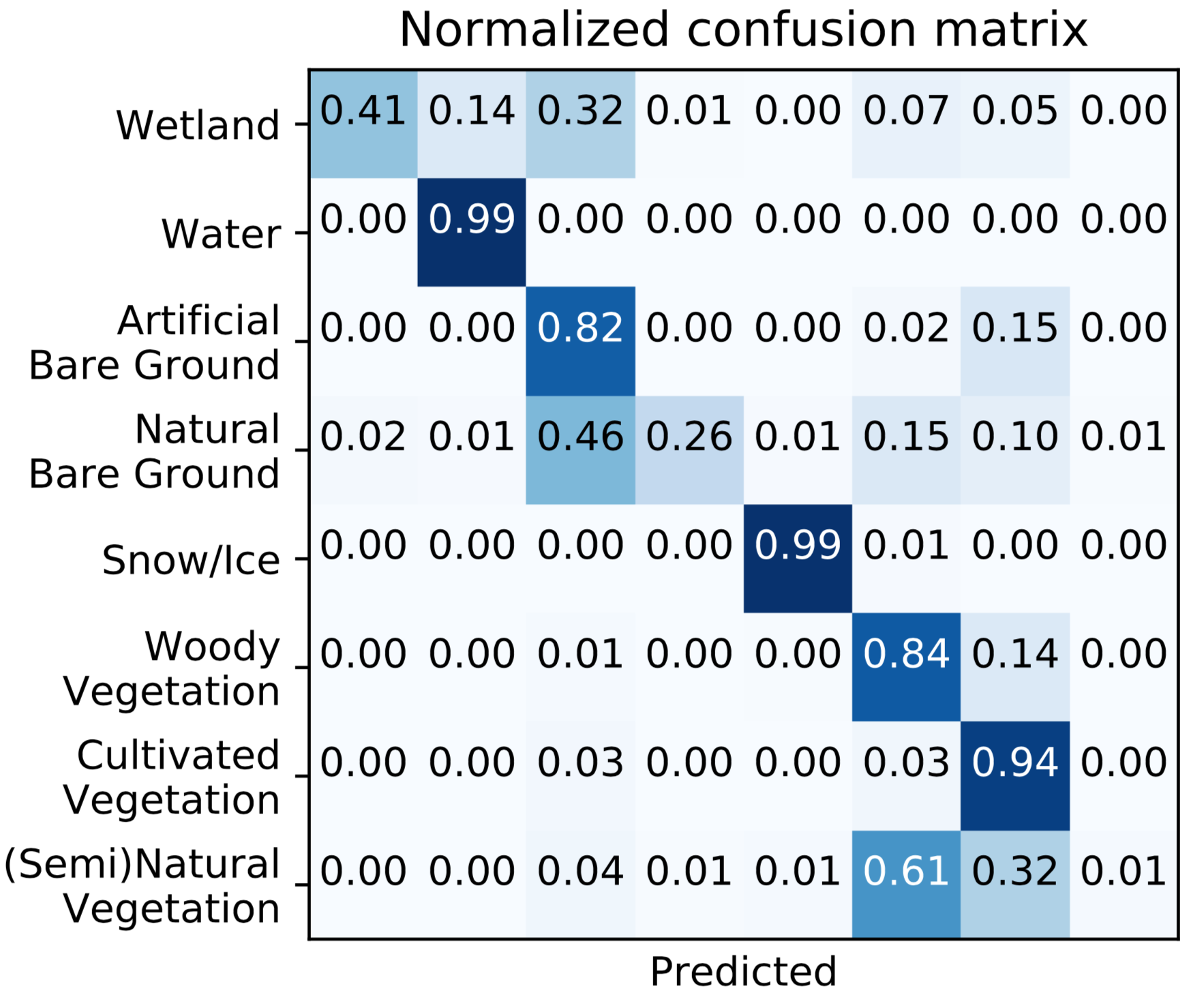}
  \caption{Normalized Confusion Matrix}
  \label{CM}
\end{figure}

\section{Results}

Our experiments included varying the number of trees for SciKit-Learn’s Random Forest Classifier, but the default of 10 estimators performed well, achieving 88.75\% average model accuracy for the scenes from the four tiles studied. Some classes, like “water” and “snow/ice”, were predicted with high accuracy and high confidence across all scenes, which is expected given their distinct spectral signatures. Other classes, like “wetland” and “(semi) natural vegetation”, are subtler and were expected to be more difficult to classify. Within vegetated classes, “woody vegetation” and “cultivated vegetation” were predicted relatively accurately and were not confused with each other, a result of including 20 m vegetation red edge bands, resampled to 10 m. “Artificial bare ground” tended to be predicted in unclassified regions (in ground truth data), taking over areas of “natural bare ground” and “cultivated vegetation” and suggesting that traces of human activity would lead to pixels classified as “artificial bare ground” in off-vegetation season.

\subsubsection*{Acknowledgments}

This study is supported by a grant from Schmidt Futures to Radiant Earth Foundation. Authors would like to thank Dr. Stanislaw Lewinski from Space Research Centre of Polish Academy of Sciences and PI of S2GLC for his recommendations. Participants of the Radiant Earth Foundation working group on “Machine Learning for Global Development“ have also contributed to this work through their extensive participation in a workshop organized on this topic in June 2018~\citep{REFML4GD}.

\bibliography{references}

\end{document}